\documentclass[twoside]{article}
\usepackage{aistats2e}
\pdfoutput=1

\usepackage{graphicx}
\usepackage{subfigure}
\usepackage{natbib}
\usepackage{amsmath,amssymb}

\newcommand{\T}{\mathcal{T}}
\newcommand{\X}{\mathcal{X}}
\newcommand{\reals}{\mathbb{R}}
\newcommand{\bigspace}{\X \times \Theta \times \reals}
\newcommand{\scrF}{\mathcal{F}}
\newcommand{\E}{\mathbb{E}}
\newcommand{\V}{\mathbb{V}}

\newcommand{\I}{\text{I}}

\newcommand{\CRM}{\text{CRM}}

\newcommand{\norm}[1]{\left | \left | #1 \right | \right |}

\newcommand{\Ber}{\text{Ber}}
\newcommand{\Be}[2]{\text{Beta} \left ( #1,#2 \right )}
\newcommand{\N}{\text{N}}

\newcommand{\Gam}{\text{Ga}}
\newcommand{\Pois}{\text{Pois}}
\newcommand{\Mult}{\text{Mult}}
\newcommand{\Dir}{\text{Dir}}
\newcommand{\Cat}{\text{Cat}}
\newcommand{\NIG}{\text{NiG}}
\newcommand{\diag}{\text{diag}}

\newcommand{\hidetext}[1]{}
\newcommand{\response}[1]{#1}

\begin{document}

%

%

\twocolumn[

\aistatstitle{A unifying representation for a class of dependent random
measures}

\aistatsauthor{ Nicholas J. Foti \And Joseph D. Futoma \And Daniel N. Rockmore
\And Sinead Williamson}

\aistatsaddress{
Dartmouth College\\Hanover, NH 03755\\
\And Dartmouth College\\Hanover, NH 03755\\
\And Dartmouth College\\Hanover, NH 03755\\
\And Carnegie Mellon University\\Pittsburgh, PA 15213\\} ]

\begin{abstract}

We present a general construction for dependent random measures based
on thinning Poisson processes on an augmented space. The framework is
not restricted to dependent versions of a specific nonparametric
model, but can be applied to all models that can be represented using
completely random measures. Several existing
dependent random measures can be seen as specific cases of this framework.
Interesting properties of the resulting measures are derived and the efficacy
of the framework is demonstrated by constructing a covariate-dependent latent 
feature model and topic model that obtain superior predictive performance.

\end{abstract}

\section{Introduction}

Motivated by a desire for flexible models that minimize assumptions
about the underlying structure of our data, Bayesian nonparametric
models have garnered much attention in the machine learning and
statistics communities.\hidetext{ For example, the Chinese restaurant process
\citep[CRP,][]{Aldous:1985} allows us
to construct mixture models where the number of components is random
and grows with the data. The Indian buffet process
\citep[IBP,][]{Griffiths:Ghahramani:2005} allows us to create
latent feature models where the cardinality of the latent feature
space is unknown a priori.} Most Bayesian nonparametric models assume observations are
exchangeable.  In real life, this assumption is usually hard to
justify. We often have side information -- for example time
stamps or geographical location -- that we believe influences the
latent structure of our data. 

There has been growing interest in models that challenge this
exchangeability assumption, while still maintaining desirable
properties of the original nonparametric processes. A dependent nonparametric process
\citep{MacEachern:1999} is defined as a distribution over collections
of measures indexed by values in some covariate space, such
that the marginal distribution at some covariate value is described by
a known nonparametric process. A number of authors have proposed
dependent versions of the Dirichlet process \citep[for example][]{Griffin:Steel:2006,
  Rao:Teh:2009, Chung:Dunson:2011, Duan:Guindani:Gelfand:2007,
  Caron:Davy:Doucet:2007}, the Pitman-Yor process
\citep{Sudderth:Jordan:2009} and the Indian buffet process
\citep{Ren:Wang:Dunson:Carin:2011, Williamson:Orbanz:Ghahramani:2010,
  Zhou:Yang:Sapiro:Dunson:Carin:2011}.

Most of the basic nonparametric processes found in the literature can be formulated in terms
of completely random measures \citep[CRMs,][]{Kingman:1967} -- distributions
over measures that assign independent masses to disjoint subsets of
the space on which they are defined. For example, the Dirichlet
process can be obtained by normalizing the CRM known as the gamma
process. The IBP can be described in terms of a mixture of Bernoulli processes,
where the mixing measure is a completely random measure known as the
beta process \citep{Thibaux:Jordan:2007}. \hidetext{Other distributions over exchangeable matrices
can be obtained in terms of CRMs in a similar way -- for example, distributions over
integer-valued matrices can be obtained by combining the gamma process
with a Poisson likelihood \citep{Titsias:2007} or the beta process with a negative
Binomial likelihood \citep{Zhou:Hannah:Dunson:Carin:2012}, and distributions over positive
real-valued matrices can be obtained by combining the gamma process
with an exponential likelihood \citep{Saeedi:BouchardCote:2011}.}

Completely random measures on some space $\Theta$ can be represented as Poisson
processes on the product space $\Theta \times \mathbb{R}^+$. In this
paper, we show that a large class of dependent nonparametric processes
can be described in terms of operations on Poisson processes on an
augmented space $\mathcal{X}\times\Theta\times\mathbb{R}^+$. This
framework offers great flexibility in the form of the dependency, and
often leads to simple posterior updates, as any conjugacy present in
the non-dependent version of the model is carried over to the dependent case. The resulting class of 
distributions contains, or is related to,
 several existing models, such as the kernel beta process
\citep{Ren:Wang:Dunson:Carin:2011} and the spatial normalized gamma
process \citep{Rao:Teh:2009}. A major contribution of this paper is to
express the relationships between these models in a simple manner, and
aid in the understanding of existing models and the development of new
models and inference techniques.

We use this framework as a basis for two models: a covariate-dependent
latent variable
model based on the beta process, and a covariate-dependent
topic model based on the gamma process. We show that incorporating
dependency can improve the predictive power of Bayesian nonparametric
models, and that by making use of
conjugacy and simpler forms of dependence, we can obtain comparable
results to existing dependent nonparametric processes, with a dramatic decrease in time spent on inference.

\section{Background}

A completely random measure (CRM) is a distribution over
measures on some measurable space $(\Theta,\mathcal{F}_\Theta)$, such that the masses
$\Gamma(A_1), \Gamma(A_2),\dots$
assigned to disjoint subsets $A_1,A_2,\dots \in \mathcal{F}_\Theta$ by a
random measure $\Gamma$ are independent \citep{Kingman:1967}. The class of completely random
measures contains important distributions such as the beta process,
the gamma process, the Poisson process and the stable subordinator. 

A CRM on $\Theta$ is characterized by a positive L\'{e}vy
measure $\nu(d\theta, d\pi)$ on the product space $\Theta\times
\reals^+$, and can be represented in terms of a Poisson process on
this space. Let 
$\Pi = \{(\theta_k, \pi_k)\}_{k=1}^\infty$ be a Poisson process
on $\Theta\times\reals^+$, with mean
measure $\nu(d\theta,d\pi)$. Then the
completely random measure with L\'{e}vy measure $\nu(d\theta,d\pi)$ can be
represented as $\Gamma = \sum_{k=1}^\infty \pi_k \delta_{(\theta_k)}$.


De Finetti's theorem tells us that any infinitely exchangeable
sequence can be described as a mixture of
i.i.d. distributions. CRMs provide the mixing
distribution in the de Finetti representation of a number of useful
exchangeable distributions. For example, we can represent the Indian
buffet process, a distribution over exchangeable binary matrices, as a
beta process mixture over countably infinite collections of
Bernoulli random variables \citep{Thibaux:Jordan:2007}. The resulting
distribution over exchangeable binary
matrices is an appropriate prior for nonparametric versions of latent
feature models. Other distributions over exchangeable
matrices have been defined using the beta process
\citep{Zhou:Hannah:Dunson:Carin:2012} and the gamma process
\citep{Titsias:2007, Saeedi:BouchardCote:2011} as mixing
measures. 

We are often interested in
learning distributions over probability distributions -- for example,
for use in clustering or density estimation. Two important
examples of such distributions -- the Dirichlet process and the
normalized stable process -- can be obtained by normalizing the gamma
process and the stable subordinator, respectively. CRMs can also be used
directly as a prior on hazard
functions in survival analysis applications \citep{Ibrahim:ChenLSinha:2005}.

\section{Construction of dependent random measures via thinned Poisson
processes}\label{sec:construction}

Let $\Pi = \{(x_i, \theta_i, \pi_i)\}_{i=1}^\infty$ be a Poisson process (PP)
on the space $\bigspace^+$.  This space has three components: $\X$, an
auxiliary space; 
$\Theta$, a space of parameter values; and $\reals^+$, which will
be the masses making up the random measures. 
Let the mean measure of $\Pi$ be described by the positive L\'{e}vy
measure $\nu(dx,d\theta,d\pi)$. While the theory herein applies for
any such L\'{e}vy measure, we will focus on the class of L\'{e}vy
measures that factorize as
\begin{equation*}
  \textstyle \nu(dx,d\theta,d\pi) = G(dx,d\theta)\nu_0(d\pi).
  \label{eqn:factorize}
\end{equation*}
This corresponds to the class of homogeneous completely random
measures, where the size of an atom is independent of its locations in
$\Theta$ and $\X$

It follows that $\Gamma = \sum_{k=1}^\infty \pi_k \delta_{(x_k,\theta_k)}$
is a CRM on $\mathcal{X}\times\Theta$.  By the mapping theorem of PPs \citep{Kingman:1993} we 
see that $B = \sum_{k=1}^\infty \pi_k \delta_{\theta_k}$
is a CRM on $\Theta$ with rate measure given by
\begin{equation*}
  \label{eqn:projrm}
    \nu_B(d\theta,d\pi) = \int_{\X} \nu(dx,d\theta,d\pi) 
    = \nu_0(d\pi) \int_{\X} G(dx,d\theta)
\end{equation*}



Let $\T$ be some covariate space -- for example time -- and let 
$\{p_x : \T \rightarrow [0,1]\}_{x\in\X}$ be a collection of functions 
indexed by $x\in\X$.  We can now construct a family of random measures $B_t$ dependent on values
$t\in\T$.  For each point $(x_k,\theta_k,\pi_k) \in \Pi$,
define a collection $\{r_k^t\}_{t\in\T}$ of Bernoulli random variables (so
$r^t_k$ is a binary valued random function on $\mathcal{T}$), such that $p(r^t_k = 1) = 
p_{x_k}(t)$.  \response{The $r^t_k$s indicate whether atom $k$ in
  the global measure $B$ appears in the local measure $B_t$ at
  covariate value $t$.  Therefore, the function $p_x$ controls the
degree of dependence between two measures $B_t$ and 
$B_{t'}$.}

Appealing to the marking theorem of PPs \citep{Kingman:1993}, we see that the 
resulting thinned PP $\Pi_t$ and its
associated rate measure $\nu_t$ are described by
\begin{align*}
  \Pi_t &= \left \{ (x_k,\theta_k,\pi_k) \; | \; r^t_k = 1 \right \}_{k=1}^\infty 
     \\
  \nu_t(A,d\theta,d\pi) &= \int_{x\in A}p_x(t)\nu(d x, d\theta, d\pi) 
\end{align*}
for $A \in \scrF_{\X}$.  Then, applying the mapping theorem to $\Pi_t$
and employing the sum form of a CRM, we find
\begin{equation*}
  \textstyle B_t = \sum_{k \; : \; r^t_k = 1} \pi_k \delta_{\theta_k} = \sum_{k = 1}^\infty
  r^t_k \pi_k \delta_{\theta_k}
  \label{eqn:dthcrm}
\end{equation*}
is a CRM on $\Theta$ that varies with $t \in \T$ and has rate measure 
\begin{equation}
  \label{eqn:dthrate}
    \nu_{B_t}(d\theta,d\pi) = \int_\X p_x(t) \nu(dx,d\theta,d\pi)
    =\nu_t(\X,d\theta,d\pi)
\end{equation}
which, given certain forms of $p_x$ and $\nu$, may be simplified
further. We refer to $\{B_t\}_{t\in\T}$ as a \emph{thinned CRM}.

If the thinning function $p_x(t)$ is taken to be a bounded unimodal kernel function 
$K(t,m,\phi)$, where $x:=(m,\phi)$ gives the center and dispersion of the
kernel, we can interpret the model as saying that each atom
$\pi_k\delta_{(x_k,\theta_k)}$ of the CRM
defined on $\mathcal{X}\times\Theta\times\reals^+$ is ``active'' in
some subregion of $\T$, dictated by a location $m_k$ and a
dispersion $\phi_k$. However, $p_x$ need not be unimodal, or even a
kernel. Later, we will consider a form for $p_x$ that allows
atoms to be active at multiple locations.

\subsection{Properties of thinned CRMs}
The moments of $B_t(A)$ for any $x \in \X$ and $A \in \mathcal{F}_{\Theta}$ can be
determined from Campbell's theorem \citep{Kingman:1993} using Eq.~\ref{eqn:dthrate}.  Another
quantity of interest is the correlation between the marginals of a
thinned CRM at two covariate values $t$ and $t'$. Assuming
that $\V(\pi_k) < \infty$, which holds for most CRMs used in practice, we have
\begin{equation*} \label{eqn:corr}
  \begin{aligned}
    &\text{Corr}(B_t(A),B_{t'}(A)) \\
    &= \frac{ \displaystyle\sum_{k:\theta_k \in A} \mathbb{E}(r^t_k
      r^{t'}_k|p^t, p^{t'}) \V(\pi_k) }{ \sqrt{
    \displaystyle\sum_{k:\theta_k \in A} \mathbb{E}((r^t_k)^2) \V(\pi_k) \displaystyle\sum_{k:\theta_k \in A} \mathbb{E}((r^{t'}_k)^2)
    \V(\pi_k)
    } } \\
    &= \frac{<p^t,p^{t'}>}{\norm{p^t}\norm{p^{t'}}}\,
  \end{aligned}
\end{equation*}
where $p^t = (p_{x_k}(t))_{k=1}^\infty$.  In other words, the correlation between the two random
measures is given by the correlation between the thinning indicators
$r^t$ and $r^{t'}$, and is independent of the L\'{e}vy measure.
\response{We can therefore
  specify the correlation between the measures at different covariate
  values through the form of $p_x$. In general, smooth functions will
  capture the intuitive notion that measures at nearby covariates
  should use a similar set of atoms. Arbitrary correlation structures
  can be obtained via appropriate choice of $p_x$.}

A key property of the construction is that the resulting dependent random
measures are of the same form as the original process -- the component $\nu_0$
of the L\'{e}vy measure that governs the atom sizes is unchanged, and
the $\pi_k$s are distributed as before.  This is desirable in order to retain conjugacy in the model being
used.  The thinned CRM framework puts very few restrictions on the original
process allowing us to construct a large family of dependent CRMs, whereas
previous constructions have been limited to specific processes.

\section{Examples}
In this section, we describe thinned CRMs with two different
dependency structures, and two hierarchical models based on such
thinned CRMs.

\subsection{A single-location thinned CRM}\label{sec:singlelabel}

One of the simplest forms of covariate dependency is to assume that
the expected correlation between two measures decreases with
increasing distance in covariate space. This can be captured by
choosing the thinning probability for each atom of the global CRM
to be a unimodal distribution centered on a point in covariate space -- as we
move away from this location in covariate space, the probability of a
covariate-dependent measure featuring this atom will decrease monotonically. Such a model is
described as:
\begin{equation} \label{eqn:uni_bplfmod}
  \begin{aligned}
    \textstyle \Gamma := \sum_{k=1}^\infty \pi_k\delta_{(x_k,\theta_k)} &\sim \CRM(\nu(dx,d\theta,d\pi))\\
    \textstyle p_x(t) &= f(|x-t|) \\
    \textstyle r_k^{t} &\sim \Ber(p_{x_k}(t))\\
    \textstyle B_t &:= \textstyle{\sum_{k=1}^\infty} r_k^t \pi_k \delta_{\theta_k} \, ,
  \end{aligned}
\end{equation}
where $\X = \T$ and $f:\mathcal{X}\rightarrow [0,1]$ is some unimodal function on $\mathcal{X}$, for 
example a scaled Gaussian density.

\subsection{A multiple-location thinned CRM}\label{sec:multlabel}

The form of dependency in Section~\ref{sec:singlelabel} is
restrictive: the probability of an atom contributing to a CRM decays
with distance in covariate space. If each atom corresponds to a
feature in a latent factor model, this means that, in practice, each
feature is only going to contribute to data points within a restricted
covariate range.

Greater flexibility can be obtained by replacing the unimodal function $f$
in Eq.~\ref{eqn:uni_bplfmod} with an arbitrary function $g$. The function
$g$ might, for example, be a Gaussian random field on $\mathcal{T}$ that has
been transformed via a sigmoid function at every value of
$t\in\mathcal{T}$. 

As a concrete example, consider one such construction of a covariate-dependent
beta process. Here, the base CRM is a homogeneous beta process, with L\'{e}vy measure
$\nu_B(dx,d\theta,d\pi) = c H(dx)B_0(d\theta)\pi^{-1} (1 -
\pi)^{c-1} d\pi$ on $\mathcal{X} \times \Theta \times [0,1]$,
for some constant $c>0$ and probability measures $H$ and $B_0$.

For the thinning function, we choose a transformed relevance vector
machine kernel. The relevance vector machine \citep{Tipping:2001:SBL} can be 
seen as the weighted sum of (a finite number of)
Gaussian kernels. Locations in the auxiliary space $\mathcal{X}$
correspond to the set of centers, weights and widths of these
kernels. A standard modeling decision, which we adopt in our
experiments, is to fix the centers of these kernels to the $L$
locations $t_1,\dots, t_L$
of the data in covariate space $\T$. Each location $x_k \in \X$ therefore
corresponds to a set of $L+1$ weights $\omega_{lk} \in \reals$, and a
(shared) width $\phi_k$ selected from a fixed dictionary $D$. Our
auxiliary space is therefore defined as $\X:=\reals^{L+1} \times D$, and
our base measure $H(dx)$ can be decomposed into a normal-inverse
Gamma prior on each of the weights, and a categorical prior on the
widths. In our experiments, 
we chose small values of $c_0$ and $d_0$ (see below) resulting in 
most $\omega_{lk}$ 
being small, which implies
that $p_{x_k}(t)$ will be large at few locations.

In order to ensure valid thinning probabilities, we
transform the RVM kernel pointwise using a probit function. The generative procedure is given by:
\begin{equation} \label{eqn:rvm_bp}
  \begin{split}
     \Gamma &:= \textstyle{\sum_{k=1}^\infty} \pi_k\delta_{(x_k,\theta_k)} 
    \sim \CRM(\nu_B(dx, d\theta, d\pi)) \\
    \omega_{lk} &\sim \NIG(0, c_0, d_0), \; \phi_k \sim \Cat(\phi_1,\ldots, \phi_D) \\
     p_{x_k}(t) &= \Phi \big( \omega_{0k} + \textstyle{\sum_{l=1}^L} \omega_{lk} \exp(-\phi_k \norm{t - t_l}_2^2) \big)\\ 
   \textstyle r_k^{t} &\sim \Ber(p_{x_k}(t))
    \\
   \textstyle B_t &:= \textstyle{\sum_{k=1}^\infty} r_k^t \pi_k \delta_{\theta_k} \, .
  \end{split}
\end{equation}


\subsection{A dependent latent feature model}\label{sec:tIBP}

We can use covariate-dependent CRMs to construct
covariate-dependent latent variable models. In such a setting, each
atom of the CRM on $\X \times \Theta \times \reals^+$ is associated
with a latent feature, and the mass of that atom parameterizes a
distribution over the weight assigned to that feature. Each data point then
selects a weight for each feature according to the masses of the atom
in the corresponding thinned CRM. 

As an example, consider a latent feature model based on the
covariate-dependent beta process described in
Eq.~\ref{eqn:rvm_bp}, where $B_0(d\theta)$ is the multivariate
Gaussian prior measure -- i.e.\ each latent feature is a real-valued
vector. For each covariate value $t\in\T$, a subset of these features, and
their corresponding atom weights $\pi_k$, are selected as in
Eq.~\ref{eqn:rvm_bp} to give a local measure $B_t = \sum_{k=1}^\infty r_k^t\pi_k\delta_{\theta_k}$. 
For each data point $n$ at covariate value $x$,
a subset of features are chosen by selecting each feature with
probability $r_k^t\pi_k$. The selected features are combined
via linear superposition, and Gaussian noise is added. The generative model is as follows:
\begin{equation*} \label{eqn:bplfmod}
  \begin{aligned}
    z_{k}^{n,t} &\sim \Ber(r_k^t \pi_k) \;,t\in\T,
    k\in\mathbb{N}, n\in \{1,\dots,N_t\}\\
    A_k &\sim \mathcal{N}(0,\sigma_{A}^2\I)\;, k\in\mathbb{N} \\
    y^{n,t} &\sim \mathcal{N}(\textstyle\sum_k z_{k}^{n,t}A_k,\sigma^2 \I),
  \end{aligned}
\end{equation*}
where $r_k^t$ and $\pi_k$ are sampled according to
Eq.~\ref{eqn:rvm_bp} and $N_t$ denotes the number of data points with covariate $t$.
In the case where each observation is associated with a unique covariate we
simplify the notation to $z_{nk}$ and $r_{nk}$.


This model is a dependent version of the linear Gaussian IBP model
proposed by \citet{Griffiths:Ghahramani:2005}. The model can be
extended by using different models for generating and combining the
features \citep{Wood:2006, Miller:2011}, or by sampling a real-valued weight $s_{k}^{n,t}$ for each
instance of a feature and combining them as  $y^{n,t} \sim
\mathcal{N}(\sum_k s^{n,t}z_{k}^{n,t}A_k,\sigma^2 \I)$
\citep{Zhou:2012:BPFA}.

\subsection{A time-varying topic model}\label{sec:topic}

Topic models are popular latent variable models that decompose a text corpus 
into the underlying topics.  Topic models define a \textit{topic} as a probability 
distribution over a finite vocabulary with $P$ terms.  The simplest topic model is 
latent Dirichlet allocation \citep[LDA,][]{Blei:Ng:Jordan:2003} where the words in each 
document are generated by first drawing which topic the word exhibits from a 
document-specific topic distribution and then drawing the actual word from the 
corresponding topic.  The basic LDA model has been extended in many ways, for example, to allow 
correlated topics \citep{Blei:Lafferty:2007,Paisley:Wang:Blei:2011}, to allow the topics to drift over 
time \citep{Blei:Lafferty:2006,Wang:Blei:2008}
and to allow topic usage to vary over time \citep{Wang:McCallum:2006}.

The thinned CRM construction described in this paper can be used to construct a time-varying topic 
model where the topics are 
assumed fixed, but the usage of the topics changes over time. 
This assumption allows the learned topics to be localized in time.  
As in \cite{Zhou:Hannah:Dunson:Carin:2012}, we formulate our topic model as a Poisson factor 
model.  \hidetext{We assume that the times at which documents are observed are considered 
known covariates, similarly to the DTM and CDTM, however, we do not impose an explicit 
Markov dependence between documents at adjacent times.  }

\hidetext{Recently, models have been formulated using the assumption that topics are constant how they are used
varies with time.  For instance the topics over time model \citep[TOT,][]{Wang:McCallum:2006} models 
topic usage over time with a beta distribution.  Using a beta distribution allows learning of topics 
that are localized int time, however, topic usage is forced to be unimodel.
The non-parametric TOT model \citep[npTOT,][]{Dubey:Hefny:Williamson:Xing:2012} extends the TOT 
model by using a Dirichlet process mixture 
to model the topic activation allowing for multi-modal appearances of topics.  A 
disadvantage of these models is that
they are forced to treat time as a random quantity in order to allow 
efficient inference.  This assumption makes sense when modeling Twitter data for 
instance where documents arrive at random times.  However, for the State of the Union corpus the 
random-timestamp assumption doesn't make sense as these addresses happen at 
predictable times.}

We use a thinned gamma process (tGaP) to model the global popularity of each 
topic and the relevance vector machine in Eq.~\ref{eqn:rvm_bp} as the thinning 
function. Let $w_{pnt}$ denote the number of times the $p$th word (in a
vocabulary of $P$ words) appears in the $n$th document at time
$t$. Let $\nu_G(dx,d\theta,d\pi)
=\nu_{G0}(d\pi)H(dx)B_0(d\theta)$, where $\nu_{G0}(d\pi) = \gamma
\pi^{-1}\exp(-\lambda \pi) d\pi)$ is the L\'{e}vy measure of the gamma
process; $B_0(d\theta)$ is the $P$-dimensional Dirichlet distribution
with parameter $\alpha_\theta$; and $H(dx)$ is the prior over
parameters for the RVM as described in Section~\ref{sec:multlabel}.

The complete model, denoted tGaP-PFA, is specified as
\begin{equation}
\label{eqn:gap_pfa}
\begin{aligned}
  \Gamma &:= \sum_{k=1}^\infty \pi_k \delta_{(x_k,\theta_k)} \sim
  \mbox{CRM}(\nu_G(dx,d\theta,d\pi))\\
  p_{x_k}(t) &= \Phi \big( \omega_{0k} + \textstyle{\sum_{l=1}^L} \omega_{lk} \exp(-\phi_k \norm{t - t_l}_2^2) \big)\\ 
   \textstyle r_k^{n,t} &\sim \Ber(p_{x_k}(t))
    \\
   \textstyle G_{n,t} &:= \textstyle{\sum_{k=1}^\infty} r_k^{n,t} \pi_k
   \delta_{\theta_k} \\
   \beta_{k}^{n,t} &\sim \Gam(e,1), n=1,\dots, N_t, k\in\mathbb{N}\\
   \tilde{w}_{pntk} &\sim \Pois(\theta_{kp} r_k^{n,t} \pi_k \beta_{k}^{n,t}) \\
    w_{pnt} &= \sum_{k=1}^\infty \tilde{w}_{pntk} \sim \Pois(\sum_{k=1}^\infty \theta_{kp} r_k^{n,t} \pi_k \beta_{k}^{n,t})
\end{aligned}
\end{equation}

where the RVM machinery is presented in Eq.~\ref{eqn:rvm_bp}.  The vectors 
$\theta_k = (\theta_{k1},\dots,\theta_{kP})$ are the topics, the atom sizes $\pi_k$ can be interpreted as the baseline 
rate that this topic generates words and the $\beta_{k}^{n,t}$ are document-specific 
modulations of the global rate so that documents can exhibit a topic more or 
less than the baseline.  

\section{Relationship with other processes}

The framework for dependent random measures proposed in
Section~\ref{sec:construction} is very general, and includes or is
related to a number of existing models, as we describe below.

\subsection{Kernel beta process}

The kernel beta process \citep[KBP,][]{Ren:Wang:Dunson:Carin:2011} has
an interesting interpretation in terms of thinned CRMs. Let $\T$ be our
covariate space, $\Psi$ be a space of possible dispersions, and define
our auxiliary space as
$\mathcal{X}:=\T\times\Psi$. Let
$\nu(dx,d\theta,d\pi):= \nu_0(d\pi)H(dx)B_0(d\theta)$ be a L\'{e}vy
measure on $\mathcal{X}\times\Theta\times\reals^+$, such that $\nu_0(d\pi) =
c\pi^{-1}(1-\pi)^{c-1}d\pi$. Let $p_x(t) = K(t,m,\psi)$ for every
$x:=m\times\psi \in \mathcal{X}$, where 
$K(\cdot,\cdot,\cdot)$ is a unimodal kernel with mean $m$ and
dispersion $\psi$ bounded above by $1$. Then, the expectation of a
realization of the corresponding thinned CRM is given by
\begin{equation*}
  \label{eqn:kbprel}
   \E_{p_x}[B_t] = \sum_{k=1}^\infty \E_{p_x}[r_k^t] \pi_k
    \delta_{\theta_k} = \sum_{k=1}^\infty K(t,m_k,\psi_k) \pi_k \delta_{\theta_k}\, ,
\end{equation*}
which is exactly the form of the KBP.  In other words, the KBP is a
mixture of kernel-thinned beta processes. The thinned beta process provides a generative
process for the KBP, which could be useful for formulating the KBP 
in a probabilistic programming language
\citep[e.g.][]{Goodman:Mansinghka:Roy:Bonawitz:Tenenbaum:2008}. In
fact, the inference algorithm described in
\citet{Ren:Wang:Dunson:Carin:2011} uses such a representation.

While the KBP can be described without using a thinned Poisson process
representation, we feel that the above derivation is easier to
interpret, and properties of the KBP can be simply derived by
appealing to well-known properties of marked Poisson processes. 
Additionally, the thinned Poisson process construction makes extending the KBP
idea to arbitrary 
CRMs simple, whereas the original construction relied on specific properties of 
the beta process.

\subsection{Spatial normalized gamma processes}

Just as a Dirichlet process can be constructed by normalizing a gamma
process, a dependent Dirichlet process can be constructed by
normalizing a thinned gamma process. If $\X = \T$ and  the thinning probability
is given by $p_{x_k}(t)
= \int_0^\infty \mathbb{I}[|t-x_k|\leq \ell] f(\ell)d\ell$ for some
distribution $f(\cdot)$ over window size $l$, then after normalization, this describes the
spatial normalized Gamma processes (SN$\Gamma$P) of
\cite{Rao:Teh:2009} where a Dirichlet process, $D_t$ exists at all covariate values.

Incorporating the Poisson process representation into the SN$\Gamma$P
model could yield a number of benefits. The inference scheme described
by \citet{Rao:Teh:2009} involves representing each $D_t$ as a mixture of
independent DPs, and performing inference in the
corresponding mixture of urn schemes. However, this
approach does not scale well to higher dimensional covariate spaces, as the number of
independent regions, and thus the number of DPs that need to be
represented, grows exponentially with the dimensionality of the space. In
addition, a naive MCMC implementation mixes poorly, necessitating the
use of expensive split/merge
Metropolis Hastings steps. Representing the SN$\Gamma$P as a
normalized thinned CRM opens up the possibility of different inference
algorithms that represent $D_t$ explicitly, and may yield more
scalable and efficient implementations. 

In addition, understanding
the model in terms of thinned Poisson processes makes it easier to
change the form of dependency from the box kernel employed in the
original paper. 
Alternative kernels,
for example an exponential kernel, could be used to give a softer
falloff and hence more flexible dependency.

\hidetext{
To perform inference in the SN$\Gamma$P, \cite{Rao:Teh:2009} divided the
covariate space $\X = \reals^d \times \reals^+$ into independent regions that
were consistent with the atom locations and window widths. This allows writing
each $G_t$ as a mixture of independent DPs, where there is a DP for each of
the independent regions of $\X$.  Inference in the model was performed with
Gibbs sampling for the case where $\X = \reals \times \reals^+$.  However, this
approach does not scale well to larger covariate spaces as the number of
independent regions, and thus the number of DPs that need to be represented
grows as $N^d$ where $N$ is the number of observed covariate locations.  This
becomes expensive as $d$ increases and makes designing inference algorithms
more challenging.  Representing the SN$\Gamma$P as a normalized thinned CRM has 
some benefits.  First, simpler inference algorithms may be constructed as we
represent $G_t$ explicitly and may be able to avoid expensive
Metropolis-Hastings steps.  Second, the sampling algorithm originally developed
for the SN$\Gamma$P is tailored to the box-kernel described above.  The thinned
CRM representation allows other types of kernels to be used, e.g. an
exponential kernel that has softer falloff than the box kernel.}

\subsection{Ornstein-Uhlenbeck Dirichlet process}

The Ornstein-Uhlenbeck Dirichlet process \citep[OUDP,][]{Griffin:2007} can be
constructed in a similar manner to the KBP, with an added
normalization step. We define $\T = \reals$ and 
$\mathcal{X}:=\reals\times\Psi$, and let
$\nu(dx,d\theta,d\pi):= \nu_0(d\pi)H(dx)B_0(d\theta)$ be a L\'{e}vy
measure on $\mathcal{X}\times\Theta\times\reals^+$ and $\Gamma$ the associated CRM. 
Let $\{G_t\}$ be
the dependent CRM obtained when we choose the
thinning probability $p_{x_k}(t) = K(t,m_k,\psi_k)$ to be an Ornstein-Uhlenbeck kernel, and 
$\nu_0(d\pi)$ to be the 
L\'{e}vy measure of a Gamma process. The OUDP is then obtained as 
$D_t(A) = \frac{\E_{p_x}[G_t(A)]}{\E_{p_x}[G_t(\Theta)]}$.
For the Ornstein-Uhlenbeck kernel, Griffin showed that $D_t$ is
a DP. Unfortunately, the proof does not easily extend to arbitrary kernels or
higher dimensional spaces.  However, treating $D_t$ as a mixture
of normalized thinned CRMs immediately shows that if we instead consider the
``complete representation'' of the random measures $\{D_t\}_{t\in\T}$,
then the marginal distributions are Dirichlet
processes,  regardless of what
kernel we pick and the covariate space.

\hidetext{\subsection{Dependent Indian buffet process}

The dependent Indian buffet process \cite{Williamson:Orbanz:Ghahramani:2010} can also be cast as in 
the thinned--CRM framework.  \textbf{Nick: Do we have enough space for this?}

\textbf{Sinead: I think we can cut a bunch of the SNGP stuff (since
  it's in our other paper)}}

\subsection{Other dependent nonparametric processes with Poisson
  process interpretations}

In addition to models that can be represented in terms of thinned
Poisson processes on an augmented space, there are a number of other
models that can be described in terms of Poisson processes.
\citet{Lin:Grimson:Fisher:2010} create a Markov chain $P_1,P_2,\dots$ of Poisson processes on
$\Theta\times\reals^+$ such that, at each time point, the
corresponding Poisson process is obtained by thinning the
previous Poisson process and superimposing an independent Poisson
process. The resulting chain of Poisson processes defines a Markov
chain of gamma processes, which are normalized to give a Markov chain
of Dirichlet processes. A similar procedure, with thinning probability
that depends on the atom sizes, underlies the size-biased
deletion form of the dependent Poyla urn model of
\cite{Caron:Davy:Doucet:2007}. These models do not fit neatly into the
framework described in this paper, and are restricted to Markovian
dependency and discrete covariate spaces.

\section{Experimental evaluation}

We illustrate the effectiveness of using a thinned CRM to relax the assumption
of exchangeability in nonparametric Bayesian modeling on both synthetic and
real data.  Specifically, we consider the approaches described in
Sections~\ref{sec:tIBP} and \ref{sec:topic}, where a probit RVM thinned CRM is
used as the basis for covariate-dependent binary latent feature models
and covariate-dependent topic models, respectively. The experiments
using binary latent feature models are performed to allow comparisons
to existing  work and to show that the proposed prior is indeed capturing the structure, and
not the likelihood. The topic model experiments are intended to
highlight the ease with which we can incorporate dependency into
more complex hierarchical models, and demonstrate the performance
gains such covariate dependency can yield.

\hidetext{We assume that we observe $N$ data points
$(x_n,y_n)$ where $x_n \in \X$ is the covariate for node $n$ and 
$y_n \in \reals^d$ is the observed feature vector.  
We use a truncated version of the
beta process for computational simplicity.  We select a truncation level $K$
and draw the masses of the beta process from a
$\Be{\frac{1}{K}}{1-\frac{1}{K}}$ distribution.
The resulting $K$-dimensional vector can be shown to converge to a draw from a
beta process as $T \rightarrow \infty$ (REF Paisley).}

\subsection{Inference}
\label{sec:rvm_inference}

Inference in the probit RVM model described in
Section~\ref{sec:multlabel} is carried out using Gibbs sampling. We consider a truncated version of the
beta process and gamma process for computational simplicity.  In both
cases, we select a truncation level $K$. To approximate the beta
process, we draw $K$ atom sizes from a
$\Be{\frac{1}{K}}{1-\frac{1}{K}}$ distribution.
The resulting $K$-dimensional vector can be shown to converge to a draw from a
beta process as $T \rightarrow \infty$
\citep{Paisley:2009:NFA}. Similarly, to approximate the gamma process,
we draw $K$ atom sizes from a $\Gam(\frac{1}{K},1)$ distribution.

The weights $\{\omega_{lk}\}$ 
can be sampled using the method
of \citet{Albert:93} using the $r_k^t$ of Eq.~\ref{eqn:rvm_bp} as
observations.  To allow conjugate updates for $\pi_k$ we introduce an auxiliary
variable $b^{n,t}_{k}$ for each data point and feature such that $z^{n,t}_{k} = 1$
iff $b^{n,t}_{k} =1$ and $r^t_{k}=1$.  
We then sample $(b^{n,t}_{k}, r^t_k)$ from their
joint distribution, which can be enumerated since both variables are binary.  A
similar scheme was used in \citet{Ren:Wang:Dunson:Carin:2011}.  Lastly, we sample 
the kernel dispersion parameters $\phi_k$ from a fixed finite dictionary of possible 
values with a uniform prior over the possible values.

For the latent feature model described in Section~\ref{sec:tIBP}, the remaining Gibbs sampling 
equations are all easily derived by conjugacy, and 
\cite{Zhou:2012:BPFA} provides most of the required distributions.
For the topic model described in Section~\ref{sec:topic}, the Gibbs
sampling equations for the remaining variables are described in the supplement.

\subsection{Dependent binary latent feature model: Synthetic data}
\begin{figure}[t!]
  \begin{center}
    \includegraphics[width=\columnwidth]{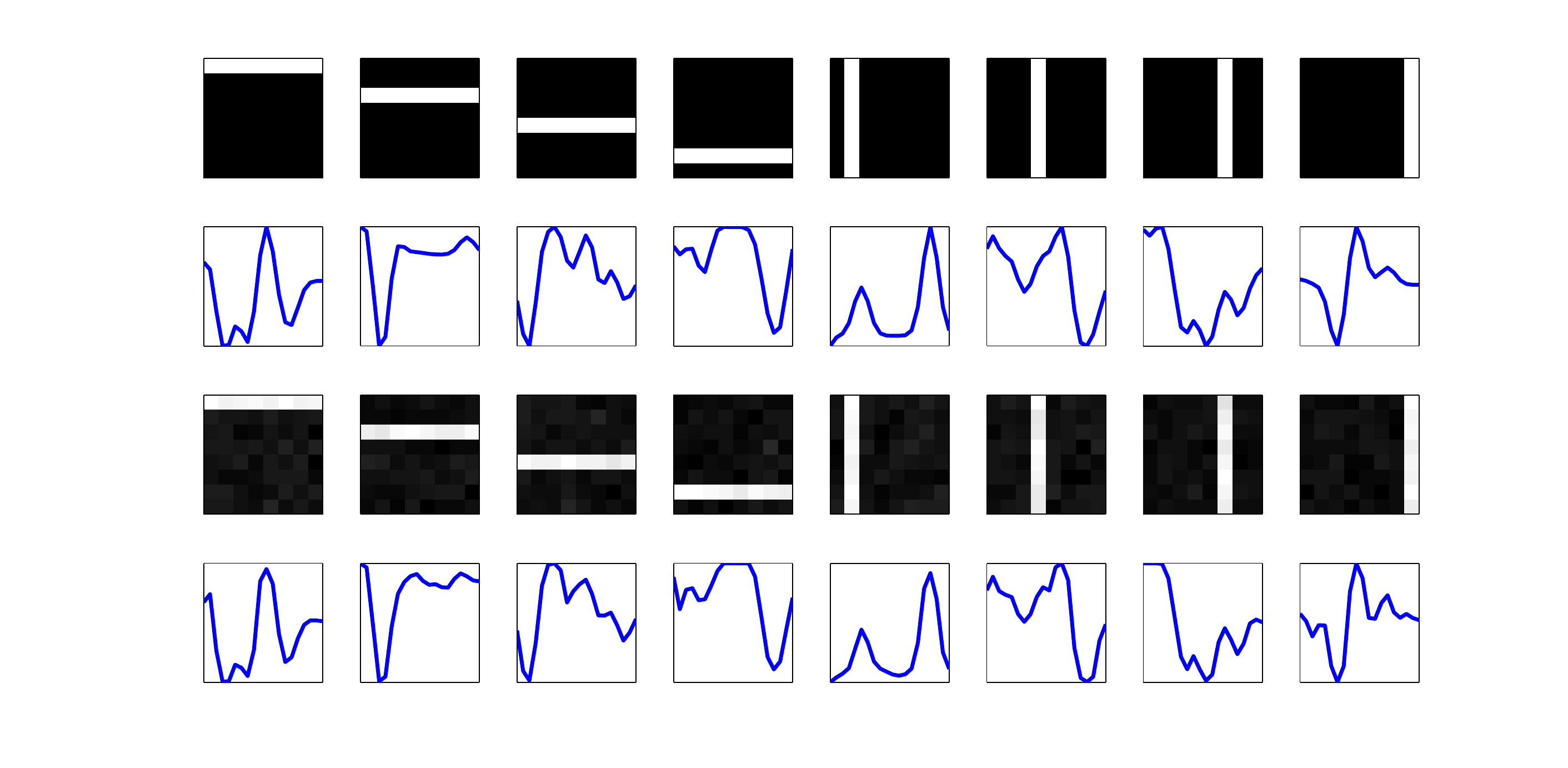}
  \end{center}
  \caption{Recovering feature probabilities in synthetic data. Top
    row: Features used to generate synthetic data. Second row:
    Time-varying thinning probabilities used to generate synthetic
    data. Third row: Recovered features (manually aligned to match
    generating features). Bottom row: Recovered time-varying thinning probabilities.}
  \label{fig:synth}
\end{figure}
To demonstrate the model's ability to uncover covariate dependent structure we
generated synthetic data similarly to the ``bag of items'' experiment
in \citet{Williamson:Orbanz:Ghahramani:2010}. Here, the covariate
space is the real line, and the covariate values are the integers $1,\ldots,20$. The data was generated 
using eight 64-pixel
image features (depicted in the top row of Fig.~\ref{fig:synth}), and eight
corresponding time-varying thinning probabilities generated using the
RVM kernel described in Section~\ref{sec:multlabel}, with kernel
weights $\omega_{lk}\sim \kappa_k \delta_0 + (1-\kappa_k)\mathcal{N}(0,4)$, 
$\kappa_k \sim \Be{1}{1}$. Each
kernel had dispersion parameter $\phi$, implying that all features vary on the
same scale.  The resulting thinning probabilities $p_{x_k}(t)$ are shown in
the second row of Fig.~\ref{fig:synth}.
For each location $t \in \{1,\ldots,20\}$ we generate a binary
matrix $Z^{y}=\{z_{n,k}^y\} \in \{0,1\}^{100 \times 8}$ of feature usage indicators for
$100$ data points at location $y$ using the sampling equation for $z^{t}_{n,k}$
in Eq.~\ref{eqn:bplfmod}.
Finally, we generate data for each $t$ as $Y^{t} = Z^{t}A + E$, where
the rows of $A$ are the eight features and  $E$ is
a matrix of observation noise with each entry normally distributed with mean
$0$ and variance $0.25$.

We perform inference using the Gibbs sampler described above, with a truncation
level of $20$ features and learned individual scales $\phi_k$ for each kernel.  
The resulting learned features and their respective
thinning probabilities are depicted in the third and fourth rows of
Fig.~\ref{fig:synth}.  The model usually learns the correct
dimensionality of the data and thinning probabilities as in the case depicted, 
however, sometimes extra
features are used to explain the noise present in the data in addition to the
correct features.  

\subsection{Dependent binary latent feature model: U.N. development indicators}
We evaluate the model in a predictive setting on a UN dataset
consisting of 15 developmental indicators for 144
countries.  This dataset was used by \citet{Williamson:Orbanz:Ghahramani:2010}
to evaluate an alternative dependent latent feature model known as the
dependent IBP (dIBP). The dIBP induces dependency directly between
corresponding elements of a collection of binary vectors using a
transformed Gaussian process. 

\response{We follow the experimental protocol used in
\citet{Williamson:Orbanz:Ghahramani:2010} where 14 countries are
selected at random as a test set, and the model is trained on the
remaining 130 countries.  For each 
test country we observe a single feature chosen at random (possibly a different
feature for each country) and the goal is to predict the remaining 14
unobserved features.}  
The covariate for
the thinned beta process model is log-GDP of the country.  We perform 10-fold
cross-validation and report the mean RMSE and two standard deviations in
Table~\ref{tab:un_cv} where we compare the results for an exchangeable
beta-Bernoulli process feature model, the thinned beta process model and the
dIBP model.  
\begin{table}[t!]
  \centering
  \caption{RMSE on UN developmental data}
  \begin{tabular}{|c|c|c|}
    \hline
    \textbf{Exchangeable} & \textbf{thinned BP} & \textbf{dIBP} \\
    \hline
    $1.02 \pm 0.08$ & $0.85 \pm 0.11$ & $0.73 \pm 0.05$ \\
    \hline
  \end{tabular}
  \label{tab:un_cv}
\end{table}

The thinned beta process model obtained lower RMSE than the
exchangeable model on all folds, indicating that incorporating
covariate information improves modeling performance. The best results
are obtained by the dIBP. This is not surprising, because the Gaussian
processes used are flexible enough to model arbitrary changes in the
latent structure. However, this added performance comes at a cost --
the dIBP uses a single Gaussian process for each latent feature, and
inference in the Gaussian processes scales cubically with the number of covariate locations. In addition,
the dIBP does not make use of conjugacy, which increases the
computational costs. While the exchangeable
model and the thinned BP models ran on the order of hours, the
dIBP ran on the order of days. We feel the thinned BP provides a
compromise between improved accuracy by taking covariate information into
account and running time.

\subsection{Time-varying topic model}

We evaluate the time-dependent topic model proposed in Section~\ref{sec:topic} both quantitatively 
and qualitatively on 
the State of the Union dataset, which consists of the full texts of the addresses 
for presidents George Washington to George Bush covering the years 1780--2002.  
As in \cite{Wang:McCallum:2006} we break up the addresses into documents of three paragraphs.  
This resulted in 5997 documents.  We created our vocabulary by computing the 
term-frequency inverse document frequency \citep[TFIDF,][]{Manning:Raghavan:Schutze:2008} score of 
all observed words and only 
keeping those with at least 10 occurrences in the corpus and in the upper 0.15 quantile of the observed 
TFIDF scores resulting in 
a vocabulary with 997 words.  All results are reported for the Dirichlet 
parameter, $\alpha_\theta = 0.05$, with comparable results obtained with other 
values.  Large values of $\alpha_\theta$ result in few topics being learned and 
vice versa.  We report the average number of topics learned for each model with with this setting of
$\alpha_\theta$ in Table~\ref{tab:perp_sotu}.

We evaluate our model using three tasks: perplexity on held out data;
time-stamp prediction; and qualitative evaluation. The perplexity evaluation was carried out
following \cite{Zhou:Hannah:Dunson:Carin:2012}, by holding out $20\%$ of the 
words from each document, training the model on the remaining $80\%$ and computing the 
perplexity of the held-out words (as described in the supplement). We
compared the tGaP-PFA model against a static version of the same model
(obtained by deterministically setting all $r_k^{n,t} = 1$) and
against the beta-negative binomial process (BNBP) model of
\cite{Zhou:Hannah:Dunson:Carin:2012}.

The perplexity results are presented in 
Table~\ref{tab:perp_sotu}. We see that the tGaP-PFA model obtains 
superior perplexity to the static version, showing that incorporating
dependency can improve performance when the data is assumed to be
non-exchangeable.  The stationary version of the tGaP-PFA model is a
much simpler model than the  BNBP topic model, which unsurprisingly
performs better. However, since the BNBP model is based on a
stationary CRM, our results suggest that a dynamic version of the BNBP topic model, constructed with a 
thinned beta process, could achieve better performance than the
stationary model.

\begin{table}[t!]
  \centering
  \caption{Average perplexity and average \# of topics for State of the Union corpus over 5 
               hold-out sets.}
  \begin{tabular}{|c|c|c|c|}
    \hline
    & \textbf{Static} & \textbf{Dynamic} & \textbf{BNBP} \\
    \hline
    \textbf{Perp.} &$624.4 \pm 2.0$ & $528.6 \pm 2.7$ & $418.1 \pm 0.9$ \\
    \hline
    \textbf{$\E[K]$} & $5.4 \pm 0.3$ & $62.6 \pm 0.8$ & $198.4 \pm 0.4$ \\
    \hline
  \end{tabular}
  \label{tab:perp_sotu}
\end{table}

We also evaluate the ability of the our dynamic model to predict the
decade of a held-out document.  We hold out $20\%$ of the documents in each decade
and train the model on the remaining $80\%$.  To predict the decade for a 
held-out document we find the decade that maximizes the predictive likelihood 
of the document.  We compare the dynamic model with a static version where we 
train a separate static tGaP-PFA model at each timestamp and predict the 
decade of a held out document by choosing the decade with the maximum 
predictive likelihood.  We also compare with a baseline prediction that selects a decade uniformly at 
random.  The results are presented in Table~\ref{tab:timestamp} where we see 
that the dynamic model obtains  $\approx 65\%$ reduction in absolute (L1) error and $\approx 4$X 
increase in accuracy over the static and baseline models.  Interestingly the 
static model performs on par with the baseline indicating substantial 
over-fitting and displaying the necessity of taking time into account.

\begin{table}[t!]
  \centering
  \caption{Predicting the decade of documents reported as absolute
    (L1) error in decades and accuracy (average over 5 hold-out sets).}
  \begin{tabular}{|c|c|c|c|}
    \hline
    & \textbf{Static} & \textbf{Dynamic} & \textbf{Baseline} \\
    \hline
    \textbf{L1} & $6.86 \pm 0.12$ & $2.42 \pm 0.02$ & $6.97 \pm 0.00$ \\
    \hline
    \textbf{Acc.} & $0.05 \pm 0.00$ & $0.21 \pm 0.01$ & $0.05 \pm 0.00$ \\

    \hline
  \end{tabular}
  \label{tab:timestamp}
\end{table}

\begin{figure}[t!]
  \begin{center}
    \includegraphics[width=\columnwidth]{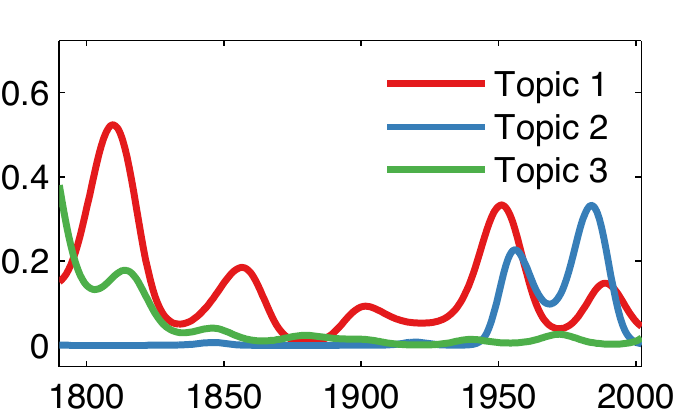}
  \end{center}
  \caption{Activation functions over time for three topics.}
  \label{fig:topic_act}
\end{figure}

\begin{table}[t!]
  \centering
  \caption{Learned topics.}
  \begin{tabular}{|c|c|c|}
    \hline
    \textbf{Topic 1} & \textbf{Topic 2} & \textbf{Topic 3} \\
    \hline
    military (0.074) & soviet (0.042) & tribes (0.142) \\
    \hline
    defense (0.061) & nations (0.035) & indian (0.124) \\
     \hline
    war (0.056) & security (0.022) & indians (0.116) \\
     \hline
    forces (0.051) & peace (0.021) & frontier (0.034) \\
    \hline
    force (0.041) & nuclear (0.020) & greater (0.027) \\
    \hline
    \end{tabular}
  \label{tab:topics}
\end{table}

In Figure~\ref{fig:topic_act} we depict the activation functions (the mean of $r^{n,t}_k$) 
over time for three topics, and in Table~\ref{tab:topics} we show the top 5 words in each topic and 
the probability of the word under the topic.  Topic 1 is a topic on war and we see it peak at 
most major conflicts that the United States was involved in.  Topic 2 is about 
the Cold War and peaks at the beginning and end.  Topic 3 regards Native 
Americans and is very prominent in addresses up to the 1850s and becomes less 
active in recent addresses.  Figure~\ref{fig:topic_act} shows that the tGaP-PFA 
topic model is able to uncover multi-modal topic activations as well as 
localizing the topic usage in time.

\section{Discussion}

We have presented a framework for dependent 
random measures, that can be used as priors for a large class of nonparametric Bayesian 
models.  Unlike previous work, our construction
is applicable to any CRM and has the added benefit that the resultant dependent 
CRMs retain any existing conjugacy.  We showed that many dependent
random measures in the
literature can actually be seen as specific cases of the thinned CRM 
framework.  We demonstrated the effectiveness of the framework by using it to
create a non-exchangeable latent feature model and a time-varying topic model.  
The models achieved superior predictive performance to exchangeable versions.

\bibliography{thincrm}
\bibliographystyle{icml2012}

\appendix

\section{Time-varying topic model}

Recall that $w_{pnt}$ represents the number of occurrences of word $p$ in 
the $n$th document at time $t$, and that we decompose this as
$w_{pnt} = \sum_{k=1}^\infty \tilde{w}_{pnt k}$, where
$\tilde{w}_{pntk}$ is the number of occurences
attributed to topic $k$.
In the generative process presented below, $p$ indexes the vocabulary, 
$t$ indexes the observed times of documents,
$n$ indexes the documents at a time $t$ and  takes values in $\{1,\ldots,N_t\}$, 
and $k$ indexes the topics.  Additionally, 
$l$ indexes the kernel functions of the RVM \citep{Tipping:2001:SBL} with 
centers $m_l$, 
which we take to be the locations of the observations (although this
is not necessary). 

The generative process is as follows
\begin{equation}
  \Gamma := \sum_{k=1}^\infty \pi_k \delta_{(x_k,\theta_k)} \sim
  \mbox{CRM}(\nu_{G0}(d\pi)H(dx)G_0(d\theta))\; ,
\end{equation}
where $x_k := (\omega_{0k},\dots, \omega_{Lk}, \phi_k)$; $\nu_{G0}(d\pi) = 
\pi^{-1}\exp(-\pi) d\pi$ is the L\'{e}vy measure of the gamma
process with parameters $(1,1)$; $B_0(d\theta)$ is the $P$-dimensional Dirichlet distribution
with parameter $\alpha_\theta$; and $H(dx) =
H_\phi(d\phi)\prod_{l=0}^LH_\omega(d \omega_l)$, where $H_\phi(d\phi)$ is the
categorical distribution over the dictionary of kernel widths, and
$H_\omega(d \omega_l) \sim \NIG(0, c_0, d_0)$ is drawn from the normal-inverse gamma distribution. 
 The rest of the model is
\begin{align}
    p_{x_k}(t) &= \Phi \big( \omega_{0k} + \textstyle{\sum_{l=1}^L} \omega_{lk} \exp(-\phi_k \norm{t - t_l}_2^2) \big)\\ 
   \textstyle r_k^{n,t} &\sim \Ber(p_{x_k}(t))
    \\
   \textstyle G_{n,t} &:= \textstyle{\sum_{k=1}^\infty} r_k^{n,t} \pi_k
   \delta_{\theta_k} \\
   \beta_{k}^{n,t} &\sim \Gam(e,1), n=1,\dots, N_t, k\in\mathbb{N}\\
   \tilde{w}_{pntk} &\sim \Pois(\theta_{kp} r_k^{n,t} \pi_k \beta_{k}^{n,t}) \\
    w_{pnt} &= \sum_{k=1}^\infty \tilde{w}_{pntk} \sim \Pois(\sum_{k=1}^\infty \theta_{kp} r_k^{n,t} \pi_k \beta_{k}^{n,t})
\end{align}

\section{Gibbs sampler}

We use a truncated version of the model by fixing the number of atoms we will 
represent to $K$ and forming the (finite) random measure, 
$\Gamma_K := \sum_{k=1}^K \pi_k \delta_{(x_k, \phi_k)}$, where $\pi_k
\sim \Gam(1/K, 1)$, $x_k:= (\omega_{0k},\dots, \omega_{Lk}, \phi_k)$,
$\omega_{lk}\sim \mbox{NiG}(0, c_0, d_0)$, and $\phi_k \sim
\{\phi_1^*,\dots, \phi_d^*\}$. 
In the limit, $K \rightarrow \infty$, $\Gamma_K \rightarrow \Gamma$ in 
distribution.  This truncation allows for the derivation of 
a straight-forward Gibbs sampler. We assume $\T$ is the set of unique
observed times.

We sample each of the variables in turn from their full conditional 
distributions.  We use a standard data-augmentation technique for probit 
regression to sample the $\omega_{lk}$ variables by introducing an auxiliary variable 
$\tilde{r}^{n,t}_{k} \sim N(p_{x_k}(t), 1)$ for each topic $k$ at each
document $n$ at
time $t$, such that
\begin{equation*}
\begin{split}
r_k^{n,t} = \begin{cases} 1 &\mbox{ if }\tilde{r}_k^{n,t} >0\\
0 &\mbox{ otherwise.}
\end{cases}
\end{split}
\end{equation*}
See \cite{Albert:93} for details of the data augmentation.
The conditional distributions are as follows.

\begin{itemize}
    \item{\textbf{Topics, $\theta_k$}}.
        \begin{equation}
            \theta_k | \ldots \sim \Dir(\alpha_\theta + \tilde{w}_{1\cdot \cdot k}, \ldots, \alpha_\theta + \tilde{w}_{P\cdot \cdot k})
        \end{equation} 
        where $\tilde{w}_{p\cdot \cdot k} = \sum_{t\in \mathcal{T}} \sum_{n=1}^{N_t} \tilde{w}_{pntk}$.
    \item{\textbf{Global topic proportions, $\pi_k$}}.
        \begin{equation}
            \pi_k | \ldots \sim \Gam(\tilde{w}_{\cdot \cdot \cdot k} + 1/K, \sum_{t\in\T}\sum_{n=1}^{N_t} \beta_{k}^{n,t} + 1)
        \end{equation}
        where $\tilde{w}_{\cdot \cdot \cdot k} = \sum_{p=1}^P \sum_{t\in\mathcal{T}} \sum_{n=1}^{N_t} \tilde{w}_{pntk}$.
    
    \item{\textbf{Per-topic counts, $\tilde{w}_{pntk}$}}.
        \begin{align}
            (\tilde{w}_{pnt1}, \dots, \tilde{w}_{pntK})  | \ldots
\notag            &\sim \Mult(w_{pnt}; \xi_{pnt1}, \ldots, \xi_{pntK}), \\
        \xi_{pntk} &= \frac{\theta_{pk} r^{n,t}_k \pi_k \beta_{k}^{n,t}}{\sum_{j=1}^K \theta_{pj} r^{n,t}_j \pi_j \beta_{j}^{n,t}}
        \end{align}
        and we ensure that the denominator is greater than $0$ by making sure 
        that when sampling the $r^{n,t}_k$s, every document is not thinning at least one topic, 
        i.e. $\forall t \forall n \exists j, r^{n,t}_j = 1$.
        
    \item{\textbf{Per-document topic rate, $\beta_{k}^{n,t}$}}.
        \begin{equation}
            \beta_{k}^{n,t} | \ldots \sim \Gam(\tilde{w}_{\cdot ntk} + a, r^{n,t}_k \pi_k + 1)
        \end{equation}
        where $\tilde{w}_{\cdot ntk} = \sum_{p=1}^P \tilde{w}_{pntk}$.
      \item{\textbf{Time-dependent indicators, $r^{n,t}_k$:}}
        There are three cases:
        \begin{enumerate}
            \item $\forall j, r^{n,t}_j = 0 \rightarrow z^{n,t}_k = 1$
            \item $\exists p, \tilde{w}_{pntk} > 0 \rightarrow r^{n,t}_k = 1$
            \item $\forall p, \tilde{w}_{pntk} = 0$ 
        \end{enumerate}
        Cases 1 and 2 are deterministic.  For case 3 
        let $u_{pntk} \sim \Pois(\rho_p)$ with $\rho_p =
        \theta_{pk}\pi_k\beta_{k}^{n,t}$  denote the fictitious 
        count of word $p$ in the $n$th document at time $t$ assigned to topic $k$ disregarding 
        $r^{n,t}_k$.  The $u_{pntk}$ allow us to determine whether $\tilde{w}_{pntk} = 0$ because the topic has been thinned or because the topic is not popular (globally or for the individual document).  
        Case 3 above then splits into the following cases:
        \begin{enumerate}
            \item $\forall p, u_{pntk} = 0, \; r^{n,t}_k = 1$ with probability $\propto p(r^{n,t}_k=1)\prod_{p=1}^P \Pois(0; \rho_p)$
            \item $\exists p, u_{pntk} > 0, \; r^{n,t}_k = 0$ with probability $\propto p(r^{n,t}_k=0) \left (1-\prod_{p=1}^P \Pois(0; \rho_p) \right )$
            \item $\forall p, u_{pntk} = 0, \; r^{n,t}_k = 0$ with probability $\propto p(r^{n,t}_k=0) \prod_{p=1}^P \Pois(0; \rho_p)$            
        \end{enumerate}
        We evaluate the three probabilities and sample from the resulting 
        discrete distribution.
    \item{\textbf{RVM weights, $\omega_{lk}$}}.  We introduce the
      auxiliary variables $\lambda_{lk}$ such that
\begin{equation*}
\begin{split}
\lambda_{lk} &\sim \Gam(c_0,d_0)\\
\omega_{lk} &\sim N(0,\lambda_{lk}^{-1})\, .
\end{split}
\end{equation*}
Let $\boldsymbol{\omega}_k = (\omega_{0k}, \ldots, \omega_{Lk})^T$
    be the vector of RVM weights and $\mathbf{\tilde{r}}_k$ be the vector of augmentation 
    variables for all all time stamps, and  
    \begin{equation}
        K_{tk} = (1, K(t, m_1, \phi_k), \ldots, K(t, m_L, \phi_k))^T
    \end{equation}
    be the vector of the evaluation of the RVM kernels for time $t$.  Then, the conditional of 
    $\boldsymbol{\omega}_k$ is given by
        \begin{equation}
            \boldsymbol{\omega}_{k} | \mathbf{\tilde{r}}_k, \ldots \sim \N(\xi, B)
        \end{equation}
        where $B = (\diag(\lambda_{0k}, \ldots, \lambda_{Lk}) + K^T_{tk} \mathbf{\tilde{r}}_k)^{-1}$ 
        and $\xi = BK^T_{tk}\mathbf{\tilde{r}}_k$.
    
    \item{\textbf{RVM auxiliary variables, $\tilde{r}^{n,t}_{k}$}}.  
    \begin{equation}
        p(\tilde{r}^{n,t}_{k} | \ldots) \propto 
            \begin{cases}
                \N(K^T_{tk} \boldsymbol{\omega}_k, 1) \mathbf{1}(\tilde{r}^{n,t}_{k} > 0),& r^{n,t}_k = 1 \\
                \N(K^T_{tk} \boldsymbol{\omega}_k, 1) \mathbf{1}(\tilde{r}^{n,t}_{k} < 0),& r^{n,t}_k = 0 
            \end{cases}
    \end{equation}
    which is a truncated normal distribution that we sample using the inversion method described in 
    \cite{Albert:93}.
    
    \item{\textbf{RVM precisions, $\lambda_{lk}$}}.
    \begin{equation}
        \lambda_{lk} | \ldots \sim \Gam \left (c_0 + \frac{1}{2}, d_0 + \frac{1}{2}\omega^2_{lk} \right )
    \end{equation}
    
    \item{\textbf{RVM kernel widths, $\phi_k$}}.
    We assume a finite dictionary $\{\phi^*_1, \ldots, \phi^*_M\}$ of possible values for the RVM kernel 
        widths, and a uniform prior on 
        these values,
        \begin{equation}
            \begin{aligned}
            &p(\phi_k = \phi^*_m | \ldots) \propto\\ 
            &\frac{1}{M} \prod_{t\in\mathcal{T}} \prod_{n=1}^{N_t} 
            \Phi(p_{\phi^*_m}(t))^{r^{n,t}_k} (1 - \Phi(p_{\phi^*_m}(t)))^{1-r^{n,t}_k}
            \end{aligned}
        \end{equation}
        where we have denoted the thinning function as a function of $\phi^*$ as the other variables 
        are held fixed.
\end{itemize}

\section{Perplexity computation}

Similarly to \cite{Zhou:Hannah:Dunson:Carin:2012}, given $B$ samples of the model parameters and 
latent variables we compute a 
Monte Carlo estimate of the held-out perplexity for unobserved counts $Y = 
[y_{p}^{n,t}]$ as 

\begin{equation}
    \begin{aligned}
    \exp \Bigg( &\frac{1}{y_{\cdot}^{\cdot, \cdot}} \sum_{p=1}^P
    \sum_{t\in\mathcal{T}} \sum_{n=1}^{N_t} y_{p}^{n,t} \times\\
    &\log \frac{\sum_{b=1}^B \sum_{k=1}^K \theta^{(b)}_{pk} \pi^{(b)}_k
      r_{ntk}{^{(b)}} \beta^{(b)}_{ntk}}
    {\sum_{b=1}^B \sum_{p=1}^P \sum_{k=1}^K \theta^{(b)}_{pk} \pi^{(b)}_k r_{ntk}{^{(b)}} \beta^{(b)}_{ntk}}
    \Bigg)
    \end{aligned}
\end{equation}
where we have used a superscript $b$ to denote the $b$th sample of the 
parameters and latent variables
and $y_{\cdot}^{\cdot, \cdot} = \sum_{p=1}^P 
\sum_{t\in\mathcal{T}}\sum_{n=1}^{N_t} y_{p}^{n,t}$ denotes the held-out number of occurrences of word $p$ in 
the $n$th document at time $t$.

\end{document}